# Reading Race: AI Recognizes Patient's Racial Identity In Medical Images

\* Banerjee I PHD[1], Bhimireddy AR MS[2], Burns JL MS[2], Celi LA MD[3,4], Chen L[5], Correa R[6], Dullerud N[7], Ghassemi M PHD[3,8], Gichoya JW MD[9], Huang S[10], Kuo P PHD[5], Lungren MP MD[10], Price BJ[11], Purkayastha S PHD[2], Pyrros AA MD[15], Oakden-Rayner L MD[12], Okechukwu C[13], Seyyed-Kalantari L PHD[14], Trivedi H MD[9], Wang R[5], Zaiman Z[6], Zhang H[7]

*Author Affiliations and addresses*

[1] Department of Biomedical Informatics, Emory University, 100 Woodruff Circle, Atlanta, GA, 30322, USA.
[2] School of Informatics and Computing, Indiana University Purdue University, 535 W Michigan St., IT 475E, Indianapolis, IN, 46202, USA.
[3] Institute for Medical Engineering and Science, Massachusetts Institute of Technology, 45 Carleton Street E25-330, Cambridge, MA, 02142, USA.
[4] Department of Medicine, Beth Israel Deaconess Medical Center, 330 Brookline Avenue, Boston, MA, 02215, USA.
[5] Department of Computer Science, National Tsing Hua University, No. 101, Section 2, Kuang-Fu Road, Hsinchu, 30013, Taiwan.
[6] Department of Computer Science, Emory University, 201 Dowman Drive, Atlanta, GA, 30322, USA.
[7] Department of Computer Science, University of Toronto, 214 College St, Toronto, ON, M5T 3A1, Canada.
[8] Department of Electrical Engineering and Computer Science, Massachusetts Institute of Technology, 77 Massachusetts Avenue, Cambridge, MA, 02139, USA.
[9] Department of Radiology, Emory University, 1364 Clifton Rd, Atlanta, GA, 30322, USA.
[10] Stanford University School of Medicine, 291 Campus Drive, Palo Alto, CA, 94305, USA.
[11] Florida State University College of Medicine, 1115 W Call St, Tallahassee, FL, 32304, USA.
[12] Australian Institute for Machine Learning, University of Adelaide, AIML Building, Lot Fourteen, Cnr North Terrace & Frome Road, Adelaide, South Australia 5000.
[13] Department of Computer Science, Georgia Institute of Technology, 266 Ferst Drive, Atlanta, GA, 30332, USA.
[14] Lunenfeld-Tanenbaum Research Institute, Sinai Health, 600 University Ave, Toronto, ON, M5G 1X5, Canada.
[15] Dupage Medical Group, 40 S Clay St, Hinsdale, IL, 60521, USA.

## Abstract

**Background:** In medical imaging, prior studies have demonstrated disparate AI performance by race, yet there is no known correlation for race on medical imaging that would be obvious to the human expert interpreting the images. We conduct a comprehensive evaluation of AI's ability to recognize patient's racial identity from medical images.

**Methods:** Using private and public datasets we evaluate: A) performance quantification of deep learning models to detect race from medical images, including the ability of these models to generalize to external environments and across multiple imaging modalities, B) assessment of possible confounding anatomic and phenotype population features, such as disease distribution and body habitus as predictors of race, and C) investigation into the underlying mechanism by which AI models can recognize race.

**Findings:** Standard deep learning models can be trained to predict race from medical images with high performance across multiple imaging modalities. Our findings hold under external validation conditions, as well as when models are optimized to perform clinically motivated tasks. We demonstrate this detection is not due to trivial proxies or imaging-related surrogate covariates for race, such as underlying disease distribution. Finally, we show that performance persists over all anatomical regions and frequency spectrum of the images suggesting that mitigation efforts will be challenging and demand further study.

---

\* Authors (Listed Alphabetically based on last name)

**Interpretation:** We emphasize that model ability to predict self-reported race is itself not the issue of importance. However, our findings that AI can trivially predict self-reported race - even from corrupted, cropped, and noised medical images - in a setting where clinical experts cannot, creates an enormous risk for all model deployments in medical imaging: if an AI model secretly used its knowledge of self-reported race to misclassify all Black patients, radiologists would not be able to tell using the same data the model has access to.

**Funding sources:**

- Drs. Gichoya and Pyrros are supported by *the National Institute of Biomedical Imaging and Bioengineering (NIBIB) MIDRC grant* of the National Institutes of Health under contracts 75N92020C00008 and 75N92020C00021.

- Funding support for Drs. Gichoya and Purkayastha was received from the US National Science Foundation #1928481 from the Division of Electrical, Communication & Cyber Systems.

- Dr. Lungren was supported by the National Library Of Medicine of the National Institutes of Health under Award Number R01LM012966.

- Dr. Celi is funded by the National Institute of Health through NIBIB R01 EB017205

- Dr Po-Chi, Li-Ching Chen and Ryan Wang are funded by the Ministry of Science and Technology, Taiwan (MOST109-2222-E-007-004-MY3).

## Introduction

Bias and discrimination in Artificial Intelligence (AI) systems has been heavily studied in the domains of language modelling[1], criminal justice [2], automated speech recognition [3] and various healthcare application domains including dermatology [4,5], mortality risk prediction [6] and healthcare utilization prediction algorithms[7] among others. In recent explorations of racial bias in AI-based facial recognition [8], we can intuitively imagine how AI models might learn to recognise different skin tones and facial features from a photo because humans have the same ability. While AI models have also been shown to produce racial disparities in the medical imaging domain [9,10], there are no known, reliable medical imaging biomarker correlates for racial identity. In other words, while it is possible to observe indications of racial identity in photographs and videos, clinical experts cannot easily identify patient race from medical images.

This lack of known visual correlation led us to question, *how is it that established AI systems can produce disparities across racial groups in medical imaging?* Given the potential for discriminatory harm in a system that is assumed to be agnostic to race, understanding how race plays a role in medical imaging models is of high importance[11]. This question is particularly timely as medical algorithms are being cleared by the FDA and other regulatory agencies[12], including systems that use medical images (such as x-ray films, CT scans, etc.) as the primary inputs [13,14]. These regulatory agencies have offered recommendations for reporting AI studies, but public regulatory documents of FDA approved companies do not show results across relevant clinical and demographic subgroups.

Race and racial identity can be difficult attributes to quantify and study in healthcare research [15], and are often incorrectly conflated with biological concepts such as genetic ancestry [16]. In this work, we define racial identity as a social, political, and legal construct that relates to the interaction between external perceptions (i.e. "how do others see me?") and self-identification, and specifically make use of the self-reported race of patients in all of our experiments.

While previous work has demonstrated the existence of racial disparities, the mechanism for these differences in medical imaging remains unexplored [17]. One prior study noted that an AI model designed to predict severity of osteoarthritis using knee X-rays could not identify the race of the patients[10], while another evaluation of chest x-rays (CXR) found that AI algorithms could predict sex, distinguish between adult and pediatric patients, and differentiate between American and Chinese patients[18]. In ophthalmology, retinal scan images have been used to predict gender, age and cardiac markers like hypertension and smoking status [19–21]. We find little prior work explicitly targeting the recognition of racial identity from medical images, likely because in clinical practice radiologists do not routinely have access to, nor rely on, demographic information like race for diagnostic tasks.

In this study, we investigate a large number of publicly and privately available large-scale medical imaging datasets and find that self-reported race is trivially predictable by AI models trained with medical image pixel data alone as model inputs. We use standard deep learning methods for each of the image analysis experiments, training a variety of common models appropriate to the tasks. First, we show that AI models are able to predict race across multiple imaging modalities, various datasets, and diverse clinical tasks. The high level of performance persists during the external validation of these models across a range of academic centers and patient populations in the United States, as well as when models are optimised to perform clinically motivated tasks. We also perform ablations that demonstrate this detection is not due to trivial proxies, such as body habitus, age, tissue density or other potential imaging confounders for race such as the underlying disease distribution in the population. Finally, we show that the features learned appear to involve all regions of the image and frequency spectrum, suggesting that mitigation efforts will be challenging.

We emphasize that the AI's ability to predict racial identity is itself not the issue of importance, but rather that this capability is trivially learned and therefore likely to be present in many medical image analysis models, providing a direct vector for the reproduction or exacerbation of the racial disparities that already exist in medical practice. This risk is compounded by the fact that human experts cannot similarly identify racial identity from medical images, meaning human oversight of AI models is of limited use to recognise and mitigate this problem. This creates an enormous risk for all model deployments in medical imaging: if an AI model relied on its ability to detect racial identity to make medical decisions, but in doing so misclassified all Black patients, clinical radiologists (who do not typically have access to racial demographic information) would not be able to tell.

## Methods:

**Datasets:**
We obtained multiple datasets, described in Table 1, covering several imaging modalities and clinical scenarios. For chest x-ray (CXR) data, we utilised MIMIC-CXR (MXR) [22], CheXpert (CXP) [23], and Emory-CXR (EMX) obtained from Emory Hospital. For limb x-ray imaging, we used the digital hand atlas (DHA) dataset [24].

For CT imaging, the model was trained on a subset of the National Lung Screening Trial (NLST) [25] dataset and externally validated on the Stanford subset of the RSNA-STR Pulmonary Embolism CT (RSPECT) Dataset[26] (we were able to separately obtain the race labels), and a CT chest dataset from Emory Hospital (EM-CT). A screening mammogram dataset (EM-Mammo) and a cervical spine x-ray dataset (EM-CS) were acquired from Emory University Hospital.

Labels describing patient races for CheXpert and RSPECT (Stanford subset) datasets collected during this project will be released publicly alongside this publication at https://aimi.stanford.edu/research/public-datasets. The information on patient subsets for non-public datasets (such as the NLST dataset) is available on request.

Ethical approval was obtained for the Emory datasets from the Emory Institutional Review Board (Chest x-ray - IRB00091978 ; Mammograms - STUDY00000673 ; Cervical hardware - IRB00111139 ; CT chest STUDY00000506). Use of the NLST dataset was approved under project NLST-782. The data in MXR has been previously de-identified, and the institutional review boards of the Massachusetts Institute of Technology (No. 0403000206) and Beth Israel Deaconess Medical Center (2001-P-001699/14) both approved the use of the database for research. The CXP and RSPECT (Stanford subset) datasets were de-identified per Stanford institutional guidelines and deemed non-human subjects research data and therefore institutional IRB was waived per policy. Research use of the data set was in compliance with the Stanford data use agreement.

*Table 1: Summary of various datasets used for race prediction experiment*

| | MXR | CXP | EMX | NLST | RSPECT (Stanford subset) | EM-CT | DHA | EM-Mammo | EM-CS |
|---|---|---|---|---|---|---|---|---|---|
| Datatype | Chest XR | Chest XR | Chest XR | Chest CT | Chest CT (PE protocol) | Chest CT | Digital Radiography XR | Breast mammograms | Lateral C-Spine XR |
| Number of patients / images | 65,079 / 371,858 | 64,740 / 223,640 | 113,818/ 227,882 | 512 / 198,475 | 270 / 72,360 | 560/ 187,513 | 691/ 691 | 27,160 / 86,669 | 980 / 10,358 |
| Sex (% female) | 48% | 41% | 47% | 36% | 53% | 51% | 49.2% | 99.7% | 49% |
| Race | Black: 19% Asian: 3% White: 68% | Black: 6% Asian: 13% White: 67% | Black: 44% Asian: 3% White: 45% | Black: 47% Asian: - White: 53% | Black: 10% Asian: - White: 90% | Black: 72% Asian: - White: 28% | Black: 48.2% White: 51.8% | Black: 51% Asian: - White: 49% | Black: 27% Asian: - White: 73% |
| Train / val / test (% of patients) | 60/10/30 | 60/10/30 | 75/12.5/12.5 | 78/10/12 | N/A | N/A | 70/10/20 | 60/20/20 | 80/10/10 |
| Sample Image | 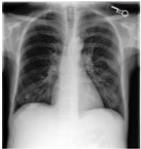 | 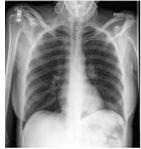 | 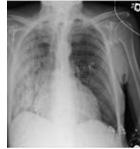 | 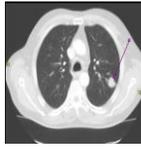 | 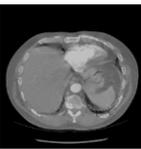 | 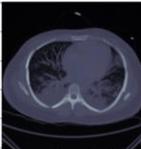 | 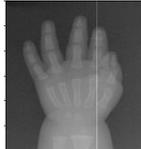 | 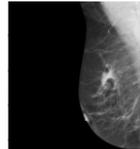 | 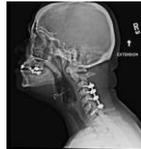 |

Investigation of possible mechanisms of race detection

To investigate the cause of previously established performance disparities by patient race, we studied several hypotheses. We performed three main groups of experiments: A) performance quantification of the deep learning models to detect race from medical images, including the ability of these models to generalize to external environments and across multiple imaging modalities, B) assessment of possible confounding anatomic and phenotype population features, such as disease distribution and body habitus as predictors of race, and C) investigation into the underlying mechanism by which AI models can recognise race. The full list of experiments is described in Table 2.

*Table 2: Summary experiments conducted to investigate mechanisms of race detection referencing the dataset used for experiment, result location (main text versus supplement) and Area Under Receiver Operating Characteristics (ROC-AUC) values for Black patients.*

|   | **Experiments** | **Dataset** | **ROC-AUC (Black)** | **Location of results** |
|---|---|---|---|---|
| A | Race detection in radiology imaging | | | |
| A1 | CXR - internal validation | MXR(Resnet34/Densenet121) CXP (Resnet 34) EMX(Resnet34/Densenet121/EfficientNet-B0) | 0.97/0.95 0.98 0.98/0.99/0.99 | Main text |
|   | CXR - external validation | MXR to CXP / MXR to EMX CXP to EMX / CXP to MXR EMX to MXR / EMX to CXP | 0.97/0.97 0.97/0.96 0.98/0.98 | Main text |
|   | CXR - comparison of models | MXR / CXP / EMX | Multiple results | Supplement |
| A2 | CT chest - internal validation | NLST (slice/study) | 0.92/0.96 | Main text |
|   | CT chest - external validation | NLST to EM-CT (slice/study) | 0.80/0.87 | Main text |
|   |   | NLST to RSPECT (slice/study) | 0.83/0.90 | Main text |
|   | Limb x-ray - internal validation | DHA | 0.91 | Main text |
|   | Mammography | EM-Mammo (image/study) | 0.82/0.84 | Main text |
|   | Cervical spine x-ray | EM-CS | 0.92 | Main text |
| A3 | CXR - models trained for | MXR - pathology detection task | 0.86 | Main text |
| | other tasks | MXR - patient reidentification task | 0.83 | Main text |
| B | Experiments on anatomic and phenotype confounders | | | |
| B1 | Body Mass Index (BMI) | CXP | 0.55/0.52 | Main text |
| | Image-based race detection stratified by BMI | EMX / MXR | Multiple results | Supplement |
| B2 | Breast density | EM-Mammo | 0.54 | Main text |
| | Breast density + Age | EM-Mammo | 0.61 | Main text |
| B3 | Disease distribution | MXR / CXP | 0.61/0.57 | Main text |
| | Image-based race detection for the "no finding" class | MXR / CXP | Multiple results | Main text |
| B4 | Removal of bone density features | MXR / CXP | 0.96/0.94 | Main text |
| B5 | Impact of Age | MXR | Multiple results | Supplement |
| | Impact of Patient sex | MXR | Multiple results | Supplement |
| C | Experiments to evaluate the mechanism of race detection | | | |
| C1 | Frequency domain filtering<br>● High-pass filtering<br>● Low-pass filtering<br>● Notch filtering<br>● Band-pass filtering | MXR<br>MXR<br>MXR<br>MXR | Multiple results<br>Multiple results<br>Multiple results<br>Multiple results | Main text<br>Main text<br>Supplement<br>Supplement |
| C2 | Image resolution and quality | MXR | Multiple results | Main text |
| C3 | Anatomical localisation | | | |
| | ● Lung segmentation experiments | MXR | Multiple results | Supplement |
| | ● Saliency maps | MXR / CXP / EMX / NLST / DHA / EM-Mammo / EM-CS | Multiple results | Supplement |
| | ● Occlusion experiments | MXR | Multiple results | Supplement |
| | ● Slice-wise results | NLST / RSPECT | Multiple results | Main text |
| C4 | Patch-based training | MXR | Multiple results | Main text |

## A: RACE DETECTION IN RADIOLOGY IMAGING

To investigate the ability of deep learning systems to discover race from radiology images, we conducted the following experiments,
1. We developed models for the detection of racial identity on three large CXR datasets with external dataset validation (A1) to establish baseline performance of AI systems for the task of race detection.
2. To evaluate whether the model performance was limited to CXRs, we trained racial identity detection models for non CXR images from multiple body locations (A2).
3. To determine if deep learning models can learn to identify racial identity when trained to perform other tasks, we trained models for pathology detection and patient re-identification and evaluated their race prediction performance (A3).

Each dataset included images, disease class labels, and race/ethnicity labels including Black/African American and White. Asian labels were available in some datasets (MXR, CXP, EMX and DHA) and were utilised when available and the population prevalence was above 1%. Hispanic/Latino labels were only available in some datasets and were coded heterogeneously, so patients with these labels were excluded from analysis.

### A1. Primary race detection

We trained and evaluated three models on the three CXR datasets (Table 1) to predict if the patient's self-reported race was Black, White or Asian. Model training details are summarized in the Supplement. Given no significant difference in performance of various architectures on race prediction, we selected the Resnet34 [27] model for external validation between the CXR datasets.

Detection performance was characterised with ROC-AUC with a one-vs-all approach for each racial group. Sensitivity and specificity are provided at the "default" operating point of 0.5, as there was no clear motivation to use a different operating point for the task.

### A2. Race detection in non-CXR imaging

Due to limited availability of the Asian class in non CXR datasets, we performed a binary classification to identify racial identity in Black patients versus White patients on the digital hand atlas, cervical spine radiographs, NLST chest CT and mammogram images (See Supplemental Table 1). The chest CT model was externally validated on the RSPECT and EM-CT datasets. Detection performance was characterised with ROC-AUC with a one-vs-all approach for each racial group. For multi slice studies, predictions were made at the slice level, with aggregated performance at the study level. Sensitivity and specificity are provided at the "default" operating point of 0.5.

A3. Race detection in models trained for other tasks

To determine if deep learning models can implicitly learn to identify race when trained to perform other detection tasks, we performed two experiments. In the first task, a Densenet-121 model was trained to detect pathology on CXR imaging using the MXR dataset while in the second task, the model was trained to re-identify unique patients in the MXR dataset. After the model was trained for the target detection task, the final classification layer was removed and the convolution weights (which were used to extract the image features) were frozen. A newly initialised softmax classification layer was trained using the hidden state from the penultimate model layer as input on the training data, which was then used to predict race on the test data. *We hypothesized that if the model was able to identify a patient's race, this would suggest the models had implicitly learned to recognize racial information despite not being directly trained for that task.*

B: EXPERIMENTS ON ANATOMICAL AND PHENOTYPE CONFOUNDERS

After establishing that high-capacity deep learning models (CNN) could identify patient race in medical imaging data, we generated a series of competing hypotheses to explain how this might occur:

1. A difference in physical characteristics between patients of different racial groups, e.g., body habitus [28] (B1) or breast density [29] (B2).
2. A difference in disease distribution amongst patients of different racial groups, e.g., Black patients have a higher incidence of certain diseases (B3) like diabetes [30], renal disease [31] and cardiac disease [32,33].
3. Location-specific or tissue-specific phenotypic or anatomical differences, e.g, blacks have higher adjusted bone mineral density than whites and a slower age-adjusted annual rate of decline in bone mineral density [34,35] (B4).
4. Cumulative effects of societal bias and environmental stress, e.g., health is generally worse for Black patients [7](B5).

B1. Race detection using body habitus

We assessed the relationship between body habitus (obtained from the recorded body mass index - BMI) and race for Black and White patients in several datasets, and with several different methods. First, we tested the correlation between BMI and race in the CXP dataset by training a we trained a generalized linear model (GLM)[36] to predict race from BMI. Secondly, we performed stratified training and testing on the MXR dataset classified into four standard BMI groups (Supplemental Table 2). Thirdly, we performed subset analysis of a trained race detection model on the EMX dataset, reporting the performance of the model at differentiating Black, White and Asian patients in four BMI groups.

B2. Tissue Density Analysis on Mammograms

We assessed the relationship between breast density and race for Black and White patients in the EM-Mammo dataset (Supplemental Table 5). We trained two distinct multi-class logistic regression models [37] (one-vs-all) to predict patient race based on the breast density and age.

B3. Race detection using disease labels

To evaluate the possibility that features related to disease distribution were responsible for the ability of models to detect race from CXRs, we trained models to predict race from the disease label data (i.e., without the images) on the MXR and CXP datasets using all available labels (14 labels, including the "no finding" and "support devices" labels).

B4. Race detection using bone density

We removed bone density information within MXR and CXP images by clipping bright pixels to 60% intensity. Sample images are shown in Figure 1. Densenet-121 models were trained on the brightness-clipped images.

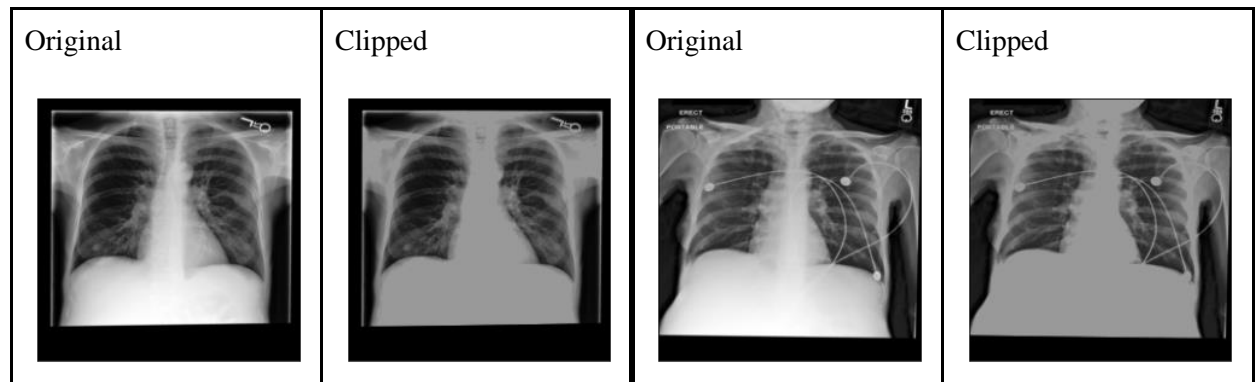

Figure 1: Samples images in MXR with bone density information removed by clipping pixels at 60% brightness.

B5. Race detection using age and sex

We investigate whether there is a cumulative effect of societal bias that impacts patients' general health, which is then used by the models as a proxy for race. We specifically examine whether there is a dose-response effect of race detection as people age, i.e., if features related to an underlying systemic health inequity are a proxy for race, then this should be more obvious in older patients. We split patients in MXR into five age groups as summarized in Supplemental Table 10 and train a Densenet121 model as described in Supplemental Table 1.

We performed a second similar experiment by splitting the MXR datasets into male and female summarized in Supplemental Table 12 and trained a Densenet121 model as described in Supplemental Table 1.

C) EXPERIMENTS TO EVALUATE THE MECHANISM OF RACE DETECTION

We investigate potential explanations of race prediction that target the known "short-cut" mechanisms high-capacity deep models may be using as proxies for race [38] by evaluating

1. Frequency domain differences in the high frequency (textural) and low frequency (structural) image features that may be predictive of race (C1).
2. How differences in image quality might influence the recognition of race in medical images, given the possibility that image acquisition practices may differ for patients with different racial identities (C2).
3. Whether specific image regions contribute to recognition of racial identity, e.g., specific patches (C3) or regional variations in the images like radiographic markers in the top right corner (C4).

C1. Frequency-domain imaging features

Given the lack of reported racial anatomical differences in the radiology literature and the known capability of deep learning models to utilise subtle textural cues that humans cannot perceive [39,40,28], we investigated the relative contributions of large-scale structural features and fine textural features by performing training and testing on datasets altered by filtering the frequency spectrum of the images in MXR. We applied low-pass filtering (LPF), high-pass filtering (HPF), bandpass filtering (BF) and notch filtering (NF) (Supplemental Figures 1 and 2 and Tables 7 and 8). After filtering, we applied the inverse Fourier transform on the filtered spectra to obtain an altered version of the original image and subsequently trained models on these perturbed datasets to observe the effect on the model's ability to predict race.

C2. Impact of image resolution and quality

To test whether race information was encoded in higher resolution images, we resized the MXR images into various resolutions and trained a Resnet34 model. To examine whether the image perturbations made an impact on race detection, we made the testing images in the MXR dataset noisy and blurred by adding gaussian noise (mean=0, variance=0.1) and applying a gaussian filter to them, respectively.

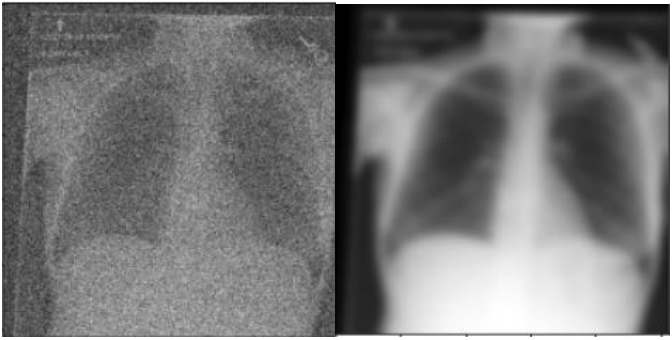
Figure 2: Examples of noisy (left) and blurred (right) images.

C3. Anatomical localisation

We investigated whether race information could be localized to a particular anatomical region or tissue by producing saliency maps for random cases for each task using the grad-cam methodology [41]. Thereafter, five radiologists performed qualitative evaluation of these artefacts. We further evaluated the significance of the regions of interest as indicated by the saliency maps by masking out the region of interest in each MXR CXR heatmap as shown in Figure 3. The masked CXRs were used to test the model trained on original MXR images.

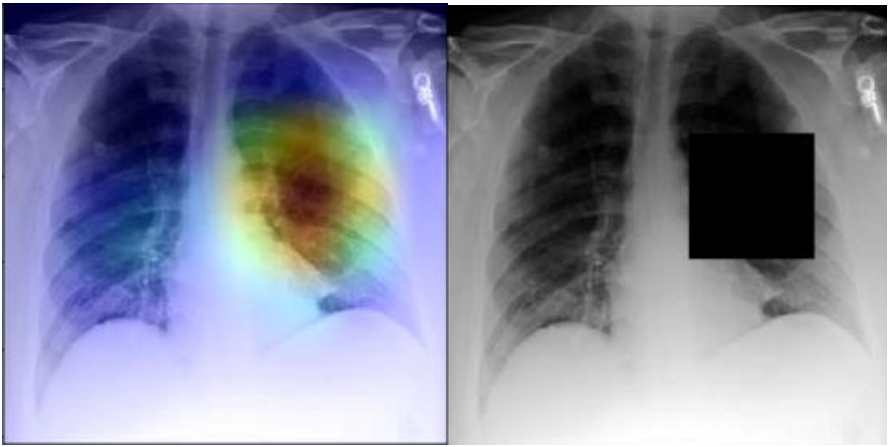
Figure 3: On the left image, there is a grad-cam saliency map showing the areas of highest probability for the race prediction model. On the right image, the pixels where the blue channels are > 0.1 are occluded with a rectangular mask.

We also tested the performance of Densenet121 using CXR images consisting of lung and non-lung segmentations using an automatic segmentation algorithm (TernausNet) [42] on the MXR dataset (Figure 4).

| Original | Non-lung segmentation | Lung segmentation |

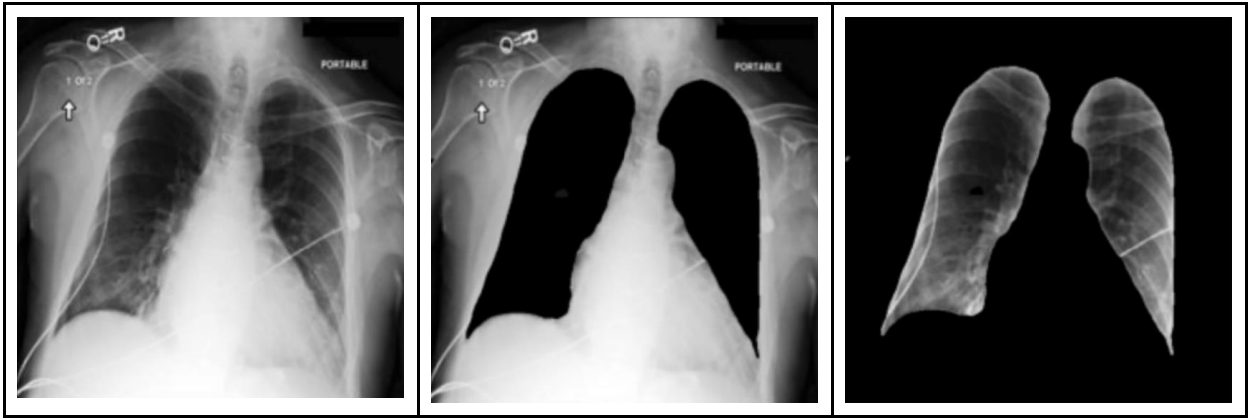

Figure 4 shows an example of lung segmentation from MXR. The original, non-lung segmented, and lung segmented images were used as the test data separately.

We analyzed slice by slice results of the CT chest model demonstrating the distribution of errors by slice-location, to reveal whether any particular anatomical region (i.e., slices from the neck, upper chest, upper abdomen etc.) appear to be more useful for race detection.

C4. Patch-based Training

We investigated whether race information can be isolated to specific patches within the chest x-ray images, for example to exclude the possibility that hospital process features such as radiographic markers were responsible for the recognition of racial identity. On the MXR dataset, we split each image into nine 3x3 square cells of equal size. We experimented with training a race prediction model using two different approaches: (1) We select one of the nine patches, and completely remove all information from the patch by setting all pixels within the patch to zero and (2) We select one of the nine patches, scale it back to the size of the original image, and use only this patch for modelling. We show an example of patched images in Figure 5. We train several networks for both approaches while varying the selected patch.

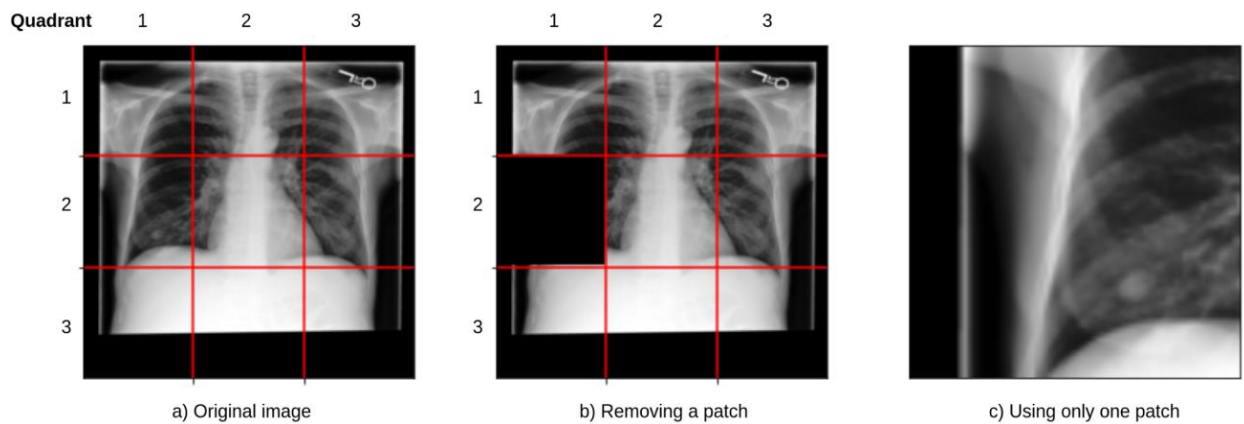

Figure 5: Sample images used for patch-based training. a) the original, unaltered image. b) the image after removing the patch located at quadrant (2, 1). c) training with only the patch located at quadrant (2, 1).

Results:

A. RACE DETECTION IN RADIOLOGY IMAGING IS EXTREMELY LEARNABLE, WITH HIGH PERFORMANCE ACROSS MODALITIES, DATASETS, TASKS, AND TRAINING OBJECTIVES

The deep learning models all demonstrated high levels of performance at the task of race detection on chest x-rays (Table 3), with sustained performance on other modalities, and strong external validations across datasets (Table 4, Table 5). Additional results comparing deep learning model architectures on the chest x-ray datasets are provided in Supplemental Table 9.

| Experiments | AUC of Race Classification | | |
| --- | --- | --- | --- |
| | Asian | Black | White |
| **A1. Primary race detection in CXR Imaging** | | | |
| MXR Resnet34 | 0.98 | 0.97 | 0.97 |
| CXP Resnet34 | 0.97 | 0.98 | 0.97 |
| EMX Resnet34 | 0.96 | 0.99 | 0.98 |
| Table 3: Performance of deep learning models for the task of race detection on three large scale chest x-ray datasets. Values reflect the area under the receiver operating characteristic curve for each model on the test set (AUC). | | | |
| **A1. External validation of race detection models in CXR imaging** | | | |
| MXR Resnet34 to CXP | 0.93 | 0.97 | 0.93 |
| MXR Resnet34 to EMX | 0.89 | 0.97 | 0.95 |
| CXP Resnet34 to MXR | 0.97 | 0.97 | 0.96 |
| CXP Resnet34 to EMX | 0.89 | 0.96 | 0.91 |
| EMX Resnet34 to MXR | 0.96 | 0.98 | 0.97 |
| EMX Resnet34 to CXP | 0.95 | 0.98 | 0.95 |
| Table 4: External validation performance of deep learning models for the task of race detection on three large scale chest x-ray datasets. Values reflect the area under the receiver operating characteristic curve for each model on the test set (AUC). | | | |

| A2. Race detection in non-CXR imaging modalities — Binary race detection (Black or White) ||
|---|---|
| NLST | 0.92 (slice), 0.96 (study) |
| NLST to EM-CT | 0.80 (slice), 0.87 (study) |
| NLST to RSPECT | 0.83 (slice), 0.90 (study) |
| EM-Mammo | 0.87 (slice), 0.91 (study) |
| EM-CS | 0.92 |
| DHA | 0.87 |

Table 5: Performance of deep learning models for the task of race detection on mammography (EM-Mammo), limb radiographs (DHA), cervical spine radiographs (EM-CS), and chest CT data (NLST, with external validation on RSPECT and EM-CT datasets). Values reflect the area under the receiver operating characteristic curve for each model on the test set (AUC) per slice and per study (by averaging the predictions across all slices).

| A3. Race detection in models trained for other task ||||
|---|---|---|---|
|  | Asian | Black | White |
| Pathology Detection MXR | 0.80 | 0.86 | 0.84 |
| Patient reidentification MXR | 0.90 | 0.83 | 0.87 |

Table 6: Performance of deep learning models for the task of race detection on CXR using models trained on the MXR dataset for the tasks of pathology detection and patient re-identification. Values reflect the area under the receiver operating characteristic curve (AUC).

## B. RACE DETECTION IS NOT DUE TO OBVIOUS ANATOMIC AND PHENOTYPE CONFOUNDER VARIABLES

### B1. Race detection using body habitus

The performance of models trained on the CXP dataset to predict race from the body mass index (BMI) alone was much lower than the image based chest x-ray models. The AUC for the generalised linear model (GLM) on the subset of patients with body mass index data was 0.55 and 0.52 on the subset of patients with a binary label corresponding to the presence or absence of BMI data.

Results for race detection on the MXR dataset stratified by body habitus are included in the Supplemental Table 3, with similar performance to A1 (AUC = [0.89, 0.98]). Subset analysis on the EMX dataset race prediction model did not show any significant performance variation from A1 across Blacks, White and Asian patients and is included in Supplemental Table 4 (AUC = [0.92, 0.99]).

### B2. Tissue density analysis on mammograms

The performance of linear regression models to classify race based on tissue density (AUC = 0.54) and on the combination of age and tissue density (AUC = 0.61) was much lower than the performance of the image models on the EM-Mammo dataset (AUC = 0.91). Additional results are summarized in Supplemental Table 6. These findings suggest breast density and age do not account for the majority of image model performance for the task of race detection.

### B3. Race detection using disease labels

As shown in Supplemental Table 18, the performance of models trained on the MXR and CXP datasets to predict race from the diagnostic labels alone was much lower than the chest x-ray image-based models with AUC values between 0.54 and 0.61 on MXR, and between 0.52 and 0.57 on CXP. Results for race detection for the "no finding" class were AUC values of 0.94, 0.94 and 0.93 for Asian, Black and White patients respectively versus 0.91, 0.95 and 0.94 for the entire dataset. These results suggest that the high AUC for racial identity recognition was not caused by disease labels.

### B4. Race detection using bone density

We find that deep learning models effectively predict patient race even when the bone density information is removed on both MXR (Black AUC = 0.96) and CXP (Black AUC = 0.94) datasets. These findings suggest that race information is not localized within the brightest pixels within the image (e.g. bone).

B5. Race detection using age and sex

Results for race detection on the MXR dataset stratified by age groups are included in Supplemental Table 11, which demonstrate no significant difference in racial identity recognition performance for patients in different age groups. There are also no differences in the AUC predictions for male and female patients in the MXR dataset (see Supplemental Table 13 for detailed results).

C. RACE INFORMATION PERSISTS IN ALL SPECTRAL RANGES AND IN THE PRESENCE OF HIGHLY DEGRADED IMAGES

C1. Race Information is present across the frequency domain

In Figure 6, we demonstrate the effect of low pass filter (LPF) and high pass filter (HPF) on model performance for various diameters in MIMIC-CXR (MXR), as well as show samples of the transformed images. We observe that LPF results in significantly degraded performance at around diameter 10, which corresponds to the level at which the images become significantly degraded visually. HPF maintains high performance up to diameter 100, which is notable as there are no visually discernible anatomical features in the sample image even at lower radius levels (i.e., performance is maintained despite extreme degradation of the image visually).

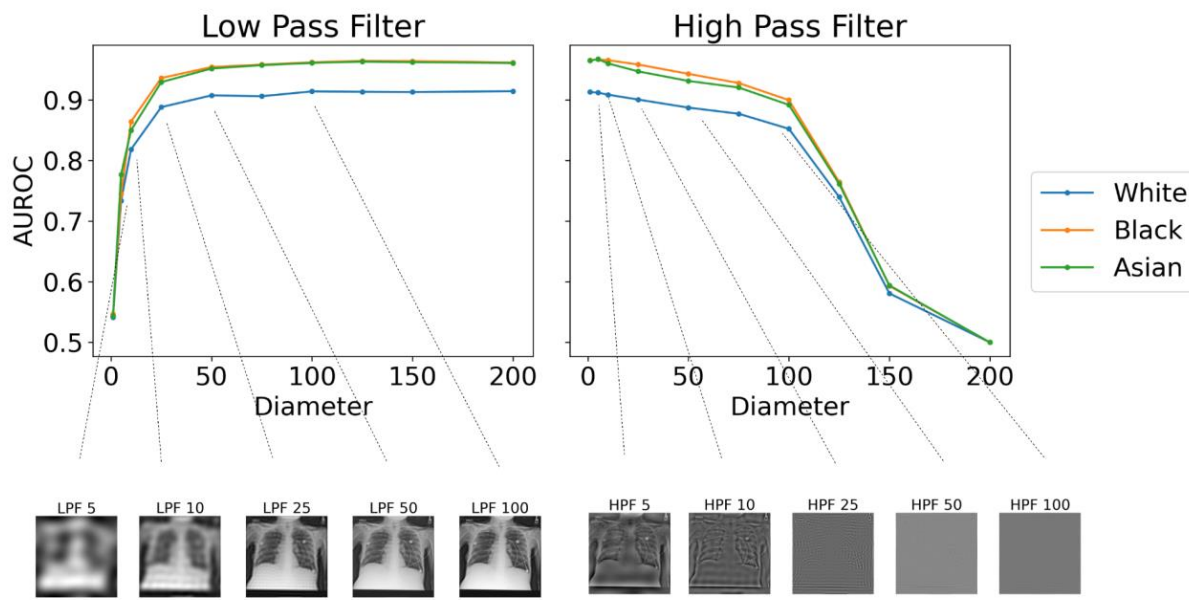

Figure 6: The performance of models trained and tested on low-pass (left) and high-pass (right) filtered images in MXR. The diameter refers to the radius of the mask used to perturb the image in the frequency domain. High AUC values are maintained across the frequency domains.

Further experiments utilising band-pass and notch filtering are reported in the Supplemental Tables 7 and 8 with the transformed images visualized in Supplemental Figures 1 and 2.

C2. Race information persists in degraded image resolution and low image quality

Figure 7 shows the AUC of various image resolutions from 4x4 resolution to 512x512 images, demonstrating AUC > 0.95 for images at 160 X 160 resolution or larger. We note a drop in performance for images below this resolution but demonstrate that race information persists more than random chance even in resolutions as small as 4x4. (See zoomed in values in Supplemental Figure 3). Similar results are observed for the perturbed images with 0.74 to 0.80 AUC for the noisy images and 0.64 to 0.72 AUC for the blurred images (Supplemental Table 15)

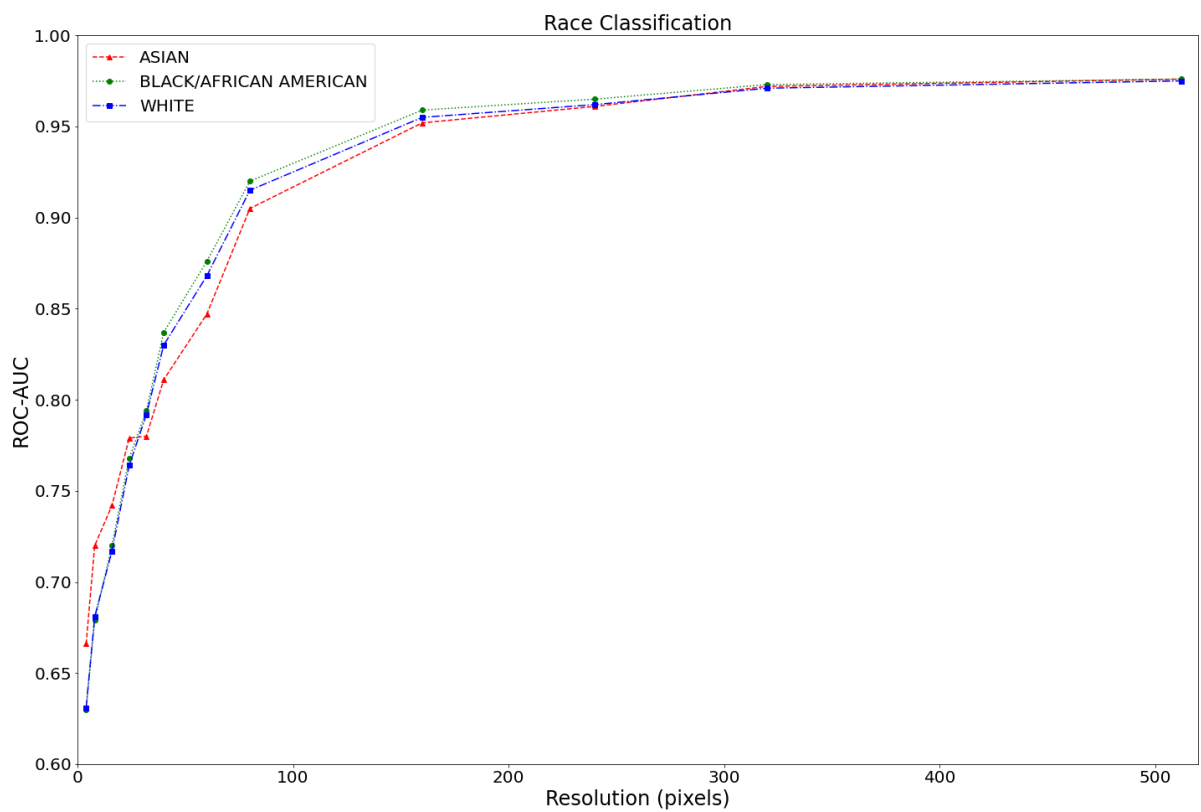

Figure 7: The performance of models trained and tested on various image resolutions from 4X4 to 512 X 512 images for Blacks, Whites and Asians. High AUC values are maintained across various image resolutions. A zoomed in plot of the lower resolutions is available in Supplemental Figure 3.

C3. Race information is not localised to a specific anatomical region or body segment.

We observe that there is no clear contribution from a specific anatomical segment using multiple experiments. There is high performance of models tested on non-lung segmentations (vs lung segmentations), but segmented predictions are lower than the original image predictions. The AUCs for testing on segmented MXR CXRs are shown in Supplemental Table 17, with higher performance on models tested on non-lung segmentations (vs lung segmentations). Race information is likely a combination of information from all lung and non lung image segments. Similar results are observed from CT slice analysis where slice-wise performance of the model is similar throughout the slices at the top, middle and lower chest. Occluding the heatmaps regions caused a slight decrease in the AUCs, but with persistent race information with AUC > 0.86 (See Supplemental Table 16)

C4. There is no clear contribution to race prediction from specific patches

In Tables 7 and 8, we present results for patch-based training. From Table 7, we observe that race prediction is very robust to the removal of any particular patch from the image, indicating that race information is not localized within a specific part of the 3x3 grid. From Table 8, we observe that there do indeed exist parts of the image with little race information (namely the lower row). However, in most cases, using only one ninth of the image is sufficient to obtain prediction performance almost on-par with using the entire image.

| Quadrant | 1 | 2 | 3 |
|---|---|---|---|
| 1 | 0.91 | 0.90 | 0.91 |
| 2 | 0.91 | 0.91 | 0.91 |
| 3 | 0.91 | 0.91 | 0.91 |

Table 7: Performance of deep learning models on race prediction for MXR (AUROC for White vs. others) when a particular patch is removed from training by setting pixel intensities to zero. Quadrants shown correspond to the geometric location of the patch (e.g. quadrant (1,1) corresponds to the top left portion of the image).

| Quadrant | 1 | 2 | 3 |
|---|---|---|---|
| 1 | 0.87 | 0.88 | 0.87 |
| 2 | 0.81 | 0.82 | 0.81 |
| 3 | 0.75 | 0.60 | 0.75 |

Table 8: Performance of deep learning models on race prediction for MXR (AUROC for White vs. others) when only one of the nine patches is used for modelling.

## Discussion

The result that deep learning models can trivially predict the self-reported race of patients from medical images alone is surprising, particularly as this task is not possible for human experts. Our work confirms that model discriminatory performance for racial identity recognition generalises across multiple different clinical environments, medical imaging modalities, and patient populations, suggesting that these models are not relying on hospital process variables or local idiosyncratic differences in how imaging studies are performed for patients with different racial identities.

While a constellation of human imperceptible signals such as scanner settings, equipment models, or image acquisition factors such as technician preferences or biases are possible explanations, our data suggests that these factors alone are insufficient to explain the discriminative performance. Beyond the demonstrated generalisability of these models to data from different clinical environments, for several of the datasets (MXR and CXP), all patients were imaged in the same locations and with the same processes independent of race. We further confirmed that the disease distribution in each dataset and the body habitus of the patients in the CXP, MXR, and EMX datasets was not strongly predictive of racial group, implying that the deep learning models were not relying on these features alone. While an aggregation of these features and interdependencies in multi-dimensional feature space may explain the ability of AI models to detect racial identity in medical images, we could not identify any specific image-based covariates that could explain the surprisingly high recognition performance presented by this work.

We investigated several hypotheses in an attempt to identify the underlying mechanism or image features responsible, including macroscopic phenotypic variation (for example, consistent anatomical differences between racial groups), regions of the images that contributed most strongly to the performance, and the presence of predictive image features which do not map directly onto any anatomical concepts.

The results of the low-pass and high-pass filtering experiments suggest that features relevant to the recognition of racial identity are present throughout the image frequency spectrum. Models trained on low-pass filtered images maintained high performance until the point of obvious image blurring. Even more dramatically, models trained on high-pass filtered images maintained performance well beyond the point that the degraded images contained no recognisable structures; to the human co-authors and radiologists it was not even clear that the image was an x-ray at all. Furthermore, the experiments involving patch based training, slice-based error analysis, and saliency mapping were non-contributory: no regions of the images consistently informed race recognition decisions. Overall, we were unable to isolate image features that are responsible for the recognition of racial identity in medical images, either by spatial location, in the frequency domain, or caused by common anatomic and phenotype confounders associated with racial identity.

One commonly proposed method to mitigate bias is through the selective removal of features that encode protected attributes such as racial identity, while retaining as much information useful for the clinical task as possible, in effect making the machine learning models "colorblind" [43]. While this approach has already been criticised as being ineffective in some circumstances [44], our work further suggests that such an approach may not succeed in medical imaging simply for the fact that racial identity information appears to be incredibly difficult to isolate. The ability to detect race was not mitigated by any reasonable reduction in resolution or by the injection of noise, nor by frequency spectrum filtering or patch based masking.

These findings suggest that not only is racial identity trivially learned by AI models, but that it appears likely that it will be remarkably difficult to debias these systems. We could only reduce the ability of the models to detect race with extreme degradation of the image quality, to the level where we would expect task performance to also be severely impaired and often well beyond that point that the images are *undiagnosable* for a human radiologist. While further research into methods to selectively reduce the presence of racial information in medical images is needed, it seems plausible that technical solutions along these lines are unlikely to succeed and that human-lead strategies designed to detect racial bias [45] and the intentional training of models to equalise racial outcomes [10] should be considered to be the default approach from a medical AI safety perspective.

The findings that models trained on CXR for other tasks, such as pathology detection and patient re-identification, will learn features predictive of self-reported race is highly concerning, although this is in keeping with previous reports on racial disparities in medical machine learning systems [4,7,9]. These results suggest that even when race is poorly correlated with the outcome of interest, as is the case in pathology detection, deep learning models are likely to learn unintended cues related to race and incorporate these cues in their decision making. It is easy to imagine many scenarios where the use of these features is not only unintended but is actively unwanted due to the risk of causing or worsening racial disparities, and it is likely that this risk has thus far been underappreciated. The regulatory environment in particular, while evolving, has not yet produced strong processes to guard against unexpected racial recognition by AI models, either to identify these capabilities in models or to mitigate the harms that may be caused.

There were several limitations to this work. Most importantly, we rely on self-reported race as the ground truth for our predictions. There has been extensive research into the correlation between self-reported race and genetic ancestry in the field of genomics, which have shown more genetic variation within races than between races, and that race is more a social than biological construct [16]. We note that in the context of racial discrimination and bias, the vector of harm is not genetic ancestry, but instead is the social and cultural construct that is racial identity, which we have defined as the combination of external perceptions and self-identification. Indeed, biased decisions are not informed by genetic ancestry information, which is not directly available to medical decision makers in almost any plausible scenario. As such, self-reported race should be considered a strong proxy for racial identity.

We were limited by the availability of racial identity labels and the small cohorts of patients from many racial identity categories. As such, we focused on Whites, Blacks and Asians, excluding patient populations which were too small to adequately analyse (for example, Native American patients) and excluding Hispanic labels due to variations in how this label was recorded across datasets.

Currently, our work has only investigated the capability of AI models to recognise racial identity in imaging studies which rely on ionising radiation (x-ray, CT). Further investigation of other modalities such as ultrasound and magnetic resonance imaging may help to isolate any underlying biological mechanisms by which racial identity information is propagated into medical images.

## Conclusion

We demonstrate that medical AI systems can easily learn to recognise racial identity in medical images, and that this capability is extremely difficult to isolate or mitigate. We strongly recommend that all developers, regulators, and users who are involved with medical image analysis consider the use of deep learning models with extreme caution. In the setting of x-ray and CT imaging data, patient racial identity is readily learnable from the image data alone, generalises to new settings, and may provide a direct mechanism to perpetuate or even worsen the racial disparities that exist in current medical practice. Our findings indicates that future medical imaging AI work should emphasize explicit model performance audits based on racial identity, sex and age, and that medical imaging datasets should include the self-reported race of patients where possible to allow for further investigation and research into the human-hidden but model-decipherable information that these images appear to contain related to racial identity.

# Supplemental material

## General model training details

**Model Architectures:**
The details of model settings are listed in the Supplemental Table 1 while datasets splits are summarized in Table 1 in the main manuscript. The CNN model architectures were selected based on dataset size and task complexity. CXR race classification models were trained using Resnet34 [1], Densenet121 [2] and EfficientNetB0 [3] architectures with pre-trained weights from ImageNet. We trained a Resnet50 [1] baseline model on the digital hand atlas, Resnet34 [1] model on cervical spine radiographs, and EfficientNetB2 [3] model on the mammogram images. A Densenet121 [2] model was trained on the NLST chest CT images and externally validated on the RSPECT and EM-CT datasets.

**Model Parameters:**
Images were resized to sizes between 224 and 320. A random seed of 2021 was used for all experiments. Hyperparameters including random horizontal flip, random 15 degree rotation, and random zoom of ±10% were applied during training. Adam optimization algorithm was chosen with a categorical cross-entropy loss function and a starting learning rate of 1e-3 that decreased by a factor of ten after two consecutive epochs without improvement in overall validation loss. We used a batch size of 256. Importantly, these experiments were performed using the standard model implementations included in the public Keras package distributed with the Tensorflow library[4], one of the most popular python libraries for CNN model development.

**Model Evaluation:**
A standardised test dataset was used across various model architectures to allow for performance comparison. Detection performance was characterised with ROC-AUC with a one-vs-all approach for each racial group. Sensitivity and specificity are provided at the "default" operating point of 0.5, as there was no clear motivation to use a different operating point for the task.

**Code availability:**
All code for the various experiments is available with an open-source license at https://github.com/Emory-HITI/AI-Vengers.

## Chest CT image preprocessing

The chest CT images were preprocessed by standardizing the rescale intercept value to 1024 and normalizing the pixel values of the images by dividing by 3000. In addition, we also dropped images that had abnormal pixel values for air. This was determined by selecting an empty patch on the image and calculating the minimum pixel value within that patch. If the value was outside of the interval [-30, 30] then the image was dropped.

## Specific model training details

### A3. Race detection in models trained for other tasks

To determine if deep learning models can implicitly learn to identify race when trained to perform other detection tasks, we performed two experiments. In the first task, a Densenet-121 model was trained to detect pathology on CXR imaging using the MXR dataset, which is labelled with 14 different pathology groups (e.g. pneumonia, cardiomegaly etc.). In the second task, a Densenet-121 model was trained to re-identify unique patients in the MXR dataset. Only patients with 20 or more frontal images in the MXR dataset were selected in the re-identification task, and there were only 1509 patients in MXR that met this criteria. For both tasks, a random dataset split of 80% training, 10% validation, 10% testing was utilized.

For this experiment, after the model was trained for the target detection task, the final classification layer was removed and the convolution weights (which were used to extract the image features) were frozen. A newly initialised softmax classification layer was trained using the hidden state from the penultimate model layer as input on the training data, which was then used to predict race on the test data. *We hypothesized that if the model was able to identify a patient's race, this would suggest the models had implicitly learned to recognize racial information despite not being directly trained for that task.*

### B1. Race detection using body habitus

We assessed the relationship between body habitus (obtained from the recorded body mass index - BMI) and race for Black and White patients in several datasets, and with several different methods.

First, we tested the correlation between BMI and race in the CXP dataset. The dataset was split into training (25,000 patients, prevalence of Black patients = 8%) and testing (7,515 patients, prevalence of Black patients = 8%) cohorts, and we trained a generalized linear model (GLM)[5] to predict race from BMI as a continuous variable (when it had been recorded) or by the presence/absence of BMI as a binary indicator variable (to exclude a racial bias in the recording of BMI).

### B2. Tissue Density Analysis on Mammograms

We assessed the relationship between breast density and race for Black and White patients in the EM-Mammo dataset (Supplemental Table 5). We trained two distinct multi-class logistic regression models [6] (one-vs-all) to predict patient race based on the breast density. First, we trained a logistic regression predictive model using python Scikit-learn *linear_model* package [7] to predict race based on the patient's recorded breast tissue density. Second, we re-trained a logistic regression model to predict race but used tissue density in addition to age as input features.

### B3. Race detection using disease labels

To evaluate the possibility that features related to disease distribution were responsible for the ability of models to detect race from CXRs, we trained models to predict race from the disease label data (i.e., without the images) on the MXR and CXP datasets using all available labels (14 labels, including the "no finding" and "support devices" labels). The disease labels for the MXR [8] and CXP [9] datasets have been published previously. We split each dataset into a 70% training, 30% test set. We trained an XGBoost[10] classifier, a L1-regularized logistic regression, and a random forest classifier to predict the patient's race. We tuned hyperparameters (maximum depth for the tree-based models, and the regularization strength of logistic regression) using 5-fold cross validation on the training set.

### C1. Frequency-domain imaging features

Given the lack of reported racial anatomical differences in the radiology literature and the known capability of deep learning models to utilise subtle textural cues that humans cannot perceive [11,12,28], we investigated the relative contributions of large-scale structural features and fine textural features by performing training and testing on datasets altered by filtering the frequency spectrum of the images.

Following the procedure outlined by previous work[11,13,14], we first transform each image into the frequency domain using a 2D Fourier transform. We then apply low-pass filtering (LPF) where we set all frequency components outside a centered circle with diameter *d* to zero, and high-pass filtering (HPF), where all frequency components within a centered circle with diameter *d* are set to zero. We also test bandpass filtering (BF) and notch filtering (NF) and report these results in Supplemental Figures 1 and 2 and Tables 7 and 8. All experiments were performed multiple times while varying the radius of the frequency spectrum filters.

After filtering, we applied the inverse Fourier transform on the filtered spectra to obtain an altered version of the original image and subsequently trained models on these perturbed datasets to observe the effect on the model's ability to predict race. These experiments were performed on the MXR dataset.

C3. Anatomical localisation

We investigated whether race information could be localized to a particular anatomical region or tissue by producing saliency maps for random cases for each task using the grad-cam methodology [15]. Thereafter, five radiologists performed qualitative evaluation of these artefacts. We used the standard keras grad-cam implementation to generate saliency maps on the chest datasets (CXP, EMX, MXR), the digital hand atlas dataset, Emory Cervical spine radiographs and mammogram datasets. For the CXR datasets, saliency maps were randomly generated from the test set for each race when correctly and incorrectly classified. The mammogram grad-cams were generated for each race and breast density classification.

We further evaluated the significance of the regions of interest as indicated by the saliency maps by masking out the region of interest in each MXR CXR heatmap as shown in Figure 3. We masked pixels with blue channels larger than 0.1 in the heatmap and then produced a minimum rectangle area to cover all masked pixels. The masked CXRs were used to test the model trained on original MXR images.

We also tested the performance of Densenet121 using CXR images consisting of lung and non-lung segmentations using an automatic segmentation algorithm (TernausNet) [16] on the MXR dataset, with manual checks on each image to exclude poorly segmented images (Figure 4). The numbers of segmented CXRs used for testing are 1001, 870, and 842 for White, Black, and Asian patients respectively. This segmentation dataset will be released through the PhysioNet (https://physionet.org/) data repository. Details of model training can be found in Supplemental Table 1.

Supplemental Table 1: Summary of model training details

| | A) Race detection in imaging | | | | | | | | |
|---|---|---|---|---|---|---|---|---|---|
| | **MODELS** | **Pretrain** | **Input size** | **Optim.** | **Loss fn** | **LR** | **Batch size** | **Epochs** | **D/Out** |
| A1 | CXR models<br>• ResNet34 | IN | 320x320 | Adam | CCE | 1e-3 | 256 | 12-16 | No |
| | • Densenet121 | IN | 224x224 | Adam | CCE | 1e-3 | 128 | 10 | 0.5 |
| | • EfficientNetB0 | IN | 224x224 | Adam | CCE | 1e-3 | 256 | 20 | 0.4 |
| A2 | CT chest | IN | 512x512 | SGD w/ mom. | BCE | 1e-4 1e-5 | 16 | ~... | |
| | Limb x-ray (IV) - DHA<br>• ResNet50 | IN | 320x320 | Adam | CCE | 1e-5 | 8 | 100 | No |
| | Mammography<br>• EfficientNetB2 | IN | 256x256 | Adam | BCE | 1e-3 | 32 | 10 (ES) | No |
| | Cervical spine x-ray<br>• ResNet34 | IN | 320x320 | Adam | CCE | 1e-3 | 64 | 12-16 (ES) | No |
| A3 | CXR - models trained for other tasks<br>• MXR - pathology detection<br>• MXR - patient reidentification task | IN | 320x320 | Adam | BCE->CCE | 1e-3 | 256 | 12-16(ES) | No |

| | B) Experiments on clinical confounders | |
|---|---|---|
| | **MODELS** | **Parameters** |
| B1 | BMI<br>• LR<br>• BMI stratified training and testing on MXR<br>• BMI subset analysis on EMX | NA<br>CXR Densenet121 as above<br><br>NA |
| B2 | Breast density<br>• LR<br>Breast density + Age<br>• LR | NA<br><br>NA |
| B3 | Disease distribution<br>• LR | Disease distribution |
|   | - RF<br><br>- XGBoost<br><br>Image-based race detection for the "no finding" class | - `LR: L1 regularization, searching C` $\in [10^{-5}, 10^{1}]$`, all other hyperparameters at default values from the scikit-learn library`<br>- `RF: 100 estimators, searching max depth` $\in \{1, 2, \cdots, 6\}$`, all other hyperparameters at default values from the scikit-learn library`<br>- `XGBoost: 100 estimators, searching max depth` $\in \{1, 2, \cdots, 6\}$`, all other hyperparameters at default values from the xgboost library`<br><br>We trained and tested the Densenet121 model using the 35,307 and 18,362 images with "no finding" labels in the MXR dataset, respectively |
| B4 | Bone density | CXR Densenet121 as above |
| B5 | Impact of Age | CXR Densenet121 as above |
|    | Impact of Patient sex | CXR Densenet121 as above |

| C) Experiments to evaluate the mechanism of race detection | | |
|---|---|---|
|   | **Experiments** | **MODELS** |
| C1 | Frequency domain imaging features | CXR Densenet121 as above |
| C2 | Image resolution and quality | - Image resolution - CXR Resnet34 as above<br>  - We resized the MIMIC-CXR (MXR) images into 512x512, 320x320, 240x240, 160x160, 80x80, 60x60, 40x40, 32x32, 24x24, 16x16, 8x8, and 4x4 resolution.<br>  - split the training, validation and testing groups by patient ID by 60%, 10%, 30% respectively.<br>- Noisy and image perturbations - CXR Densenet121 as above |
| C3 | Anatomical localisation<br><br>- Lung segmentation experiments<br>  - Seg model<br>  - Classification model<br><br>- Saliency maps<br><br>- Occlusion experiments<br><br>- Slice-wise results | The numbers of segmented CXRs used for testing are 1001, 870, and 842 for White, Black, and Asian patients respectively<br>TernausNet.<br>As above...<br><br><br>Grad cam (keras)<br><br>CXR Densenet121 as above<br><br>N/A |

| | C4 | Patch-based training | CXR Densenet121 as above |
|---|---|---|---|

Supplemental Table 2: Table showing data distribution across four BMI groups (Underweight: <18.5, Normal: 18.5 to < 25, Overweight: 25 to < 30 and Obese > 30) in the MXR dataset showing the train/test split for experiment B1.

| | Obese (BMI > 30) | | Overweight (BMI 25 to < 30) | | Normal (BMI 18.5 to < 25) | | Underweight (BMI <18.5) | |
|---|---|---|---|---|---|---|---|---|
| | Train | Test | Train | Test | Train | Test | Train | Test |
| White | 5262 | 2895 | 5591 | 2498 | 5392 | 2655 | 636 | 251 |
| | 25.8% | 28.8% | 27.5% | 24.9% | 26.5% | 26.5% | 3.1% | 2.5% |
| Black | 974 | 562 | 815 | 302 | 746 | 381 | 134 | 128 |
| | 4.8% | 5.6% | 4% | 3% | 3.7% | 3.8% | 0.7% | 1.3% |
| Asian | 57 | 33 | 444 | 168 | 256 | 156 | 50 | 8 |
| | 0.3% | 0.3% | 2.2% | 1.7% | 1.3% | 1.6% | 0.2% | 0.08% |

Note: Chi-square test implies that the two factors (BMI and Race) are not independent ($p < 0.05$).

Supplemental Table 3: Table showing the AUC of race detection in four BMI groups after stratified training and testing of a Densenet121 model on the MXR dataset using data splits in Supplemental Table 2.

| | Obese (BMI > 30) | Overweight (BMI 25 to < 30) | Normal (BMI 18.5 to < 25) | Underweight (BMI <18.5) |
|---|---|---|---|---|
| White | 0.92 | 0.93 | 0.90 | 0.96 |
| Black | 0.93 | 0.96 | 0.89 | 0.97 |
| Asian | 0.91 | 0.92 | 0.94 | 0.98 |

Supplemental Table 4: showing the results of subset analysis of a trained race detection model on the Emory CXR (EMX) dataset and race AUCs in four different BMI categories.

| | Obese (BMI > 30) | Overweight (BMI 25 to < 30) | Normal (BMI 18.5 to < 25) | Underweight (BMI <18.5) |
|---|---|---|---|---|
| White | 0.99 | 0.98 | 0.97 | 0.97 |
| Black | 0.99 | 0.99 | 0.98 | 0.98 |
| Asian | 0.94 | 0.96 | 0.93 | 0.92 |

Supplemental Table 5: Table showing data distribution across four breast density groups in the EM-Mammo dataset showing the train/test split for experiment B2. The dataset was split into training (16,296 patients), validation (5,432 patients), and testing (5,432 patients). Four groups of breast density were available in the dataset - (1 - fatty, 2 - scattered fibroglandular density, 3 - heterogeneously dense and 4 - extremely dense breasts). Most patients have scattered and heterogeneous breast density. There was no statistically significant difference across the racial subgroups.

|  | 1 (Fatty) | | 2 (Scattered) | | 3 (Heterogeneous) | | 4 (Dense) | |
| --- | --- | --- | --- | --- | --- | --- | --- | --- |
|  | Train | Test | Train | Test | Train | Test | Train | Test |
| White | 2,594 (3%) | 888 (1.0%) | 13,007 (15.1%) | 4,345 (5.0%) | 14,758 (17.1%) | 4,798 (5.5%) | 1,800 (2.1%) | 615 (0.7%) |
| Black | 4,341 (5%) | 1,405 (1.6%) | 15,000 (17.3%) | 5,003 (5.7%) | 11,936 (13.8%) | 4,061 (4.7%) | 1,298 (1.5%) | 463 (0.5%) |

Note: Chi-square test implies that the two factors (Density and Race) are not independent ($p < 0.05$).

Supplemental Table 6: Table showing slice and study AUC values of race prediction across four breast density classes, and the overall dataset not split by breast density. There is no significant difference between the AUC values at various densities, and also between slice versus study AUC values.

| Tissue Density | ROC AUC (Slice) | ROC AUC (Study) |
| --- | --- | --- |
| 1 (Fatty) | 0.79 | 0.81 |
| 2 (Scattered) | 0.82 | 0.84 |
| 3 (Heterogeneous) | 0.83 | 0.84 |
| 4 (Dense) | 0.74 | 0.76 |
| Overall | 0.82 | 0.84 |

Supplemental Figure 1 showing transformed MXR images after bandpass filtering using various values of $d_1$ and $d_2$

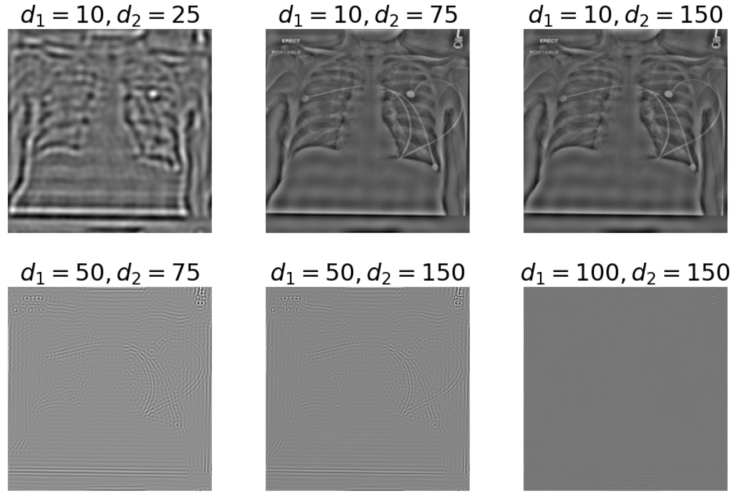

Supplemental Table 7 showing race detection performance (as AUROC for white patients) using bandpass filtering on MXR for various values of $d_1$ and $d_2$. We observe that race information is present on all examples of transformed images even when barely perceptible to the human as a CXR.

| $d_1$ \| $d_2$ | 25 | 50 | 75 | 100 | 125 | 150 |
|---|---|---|---|---|---|---|
| **10** | 0.86 | 0.90 | 0.91 | 0.91 | 0.91 | 0.91 |
| **25** |  | 0.86 | 0.89 | 0.90 | 0.90 | 0.91 |
| **50** |  |  | 0.87 | 0.89 | 0.89 | 0.89 |
| **75** |  |  |  | 0.85 | 0.86 | 0.87 |
| **100** |  |  |  |  | 0.84 | 0.84 |
| **125** |  |  |  |  |  | 0.75 |

Supplemental Figure 2 showing transformed MXR images after notch filtering using various values of $d_1$ and $d_2$

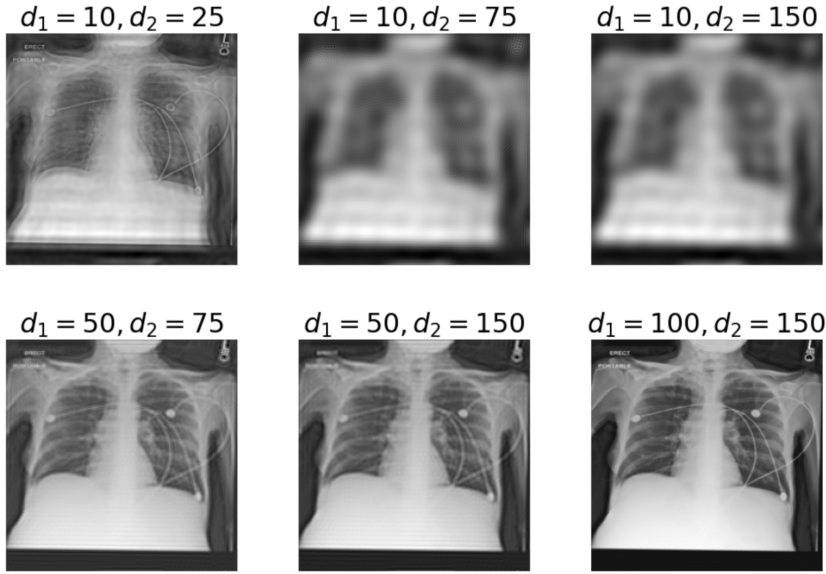

Supplemental Table 8 showing race detection performance (as AUROC for white patients) using notch filtering on MXR for various values of $d_1$ and $d_2$. We observe that race information is present on all examples of transformed images even when barely perceptible to the human as a CXR.

| $d_1$ \| $d_2$ | 25 | 50 | 75 | 100 | 125 | 150 |
|---|---|---|---|---|---|---|
| 10 | 0.90 | 0.89 | 0.87 | 0.85 | 0.82 | 0.82 |
| 25 |  | 0.90 | 0.90 | 0.89 | 0.89 | 0.89 |
| 50 |  |  | 0.91 | 0.91 | 0.91 | 0.90 |
| 75 |  |  |  | 0.91 | 0.91 | 0.91 |
| 100 |  |  |  |  | 0.91 | 0.91 |
| 125 |  |  |  |  |  | 0.91 |

Supplemental Table 9 showing comparative predictions using multiple architectures for primary race prediction on the MXR, EMX and CXP datasets. High AUCs are observed for Whites, Blacks and Asians across the three model architectures - Resnet34, Densenet121 and EfficientNetB0.

| Experiments | AUC of Race Classification | | |
|---|---|---|---|
|  | Asian | Black | White |
| MXR Densenet121 | 0.93 | 0.94 | 0.94 |
| CXP Resnet34 | 0.97 | 0.98 | 0.97 |
| EMX Densenet121 | 0.90 | 0.96 | 0.95 |
| EMX EfficientNet-B0 | 0.95 | 0.99 | 0.98 |

| | | | | | | | | | | |
|---|---|---|---|---|---|---|---|---|---|---|
| EMX Resnet34 | | | | | | | 0.96 | 0.99 | | 0.98 |

Supplemental Table 10 showing data distribution across five age groups in the MXR dataset including the train/test split for experiment B5.

| Age (yrs) | 0-20 | | 20-40 | | 40-60 | | 60-80 | | 80+ | |
|---|---|---|---|---|---|---|---|---|---|---|
| | Train | Test | Train | Test | Train | Test | Train | Test | Train | Test |
| White | 269 | 173 | 7,085 | 4,082 | 25,312 | 12,887 | 39,938 | 20,439 | 17,229 | 8,181 |
| | 0.2% | 0.3% | 6.1% | 7% | 21.7% | 22% | 34.3% | 34.9% | 14.8% | 14% |
| Black | 100 | 43 | 3,122 | 1,796 | 8,170 | 3,926 | 8,513 | 3,794 | 2,151 | 1,073 |
| | 0.09% | 0.07% | 2.7% | 3.1% | 7% | 6.7% | 7.3% | 6.5% | 1.8% | 1.8% |
| Asian | 20 | 20 | 491 | 279 | 1,176 | 657 | 1,986 | 947 | 843 | 318 |
| | 0.02% | 0.03% | 0.4% | 0.5% | 1% | 1.1% | 1.7% | 1.6% | 0.7% | 0.5% |

Note: Chi-square test implies that the two factors (Age and Race) are not independent ($p < 0.05$).

Supplemental Table 11 shows the AUC of race detection in each age group after training a Densenet121 model on the MXR dataset. The low prediction value on the 0-20 age group for the Asian class is likely due to the small dataset size which is <1%.

| | 0-20 | 20-40 | 40-60 | 60-80 | 80+ |
|---|---|---|---|---|---|
| White | 0.91 | 0.90 | 0.93 | 0.95 | 0.92 |
| Black | 0.95 | 0.91 | 0.94 | 0.95 | 0.93 |
| Asian | 0.85 | 0.92 | 0.94 | 0.96 | 0.93 |

Supplemental Table 12 showing data distribution for male and female groups in the MXR dataset including the train/test split for experiment B5.

| | Male | | Female | |
|---|---|---|---|---|
| | Train | Test | Train | Test |
| White | 50,765 | 25,378 | 39,068 | 20,384 |
| | 43.6% | 43.3% | 33.6% | 34.8% |
| Black | 9,244 | 4,177 | 12,832 | 6,455 |
| | 7.9% | 7.1% | 11% | 11% |
| Asian | 2,580 | 1,149 | 1,936 | 1,072 |
| | 2.2% | 2% | 1.7% | 1.8% |

Note: Chi-square test implies that the two factors (Sex and Race) are not independent ($p < 0.05$).

Supplemental Table 13 shows the AUC of race detection for males and females after training a Densenet121 model on the MXR dataset described in Supplemental Table 12 .

| | Asian | Black | White |
|---|---|---|---|
| MXR Densenet121 - Original | 0.93 | 0.94 | 0.94 |
| MXR Densenet121-Male | 0.94 | 0.92 | 0.91 |
| MXR Densenet121-Female | 0.95 | 0.95 | 0.95 |

Supplemental Figure 3: Zoomed plots of AUC predictions at lower image resolutions and corresponding appearance of CXR images. Lower resolution images like 4X4 have no humanly perceptible image that a radiologist would describe as a CXR, yet show AUC values > 0.60.

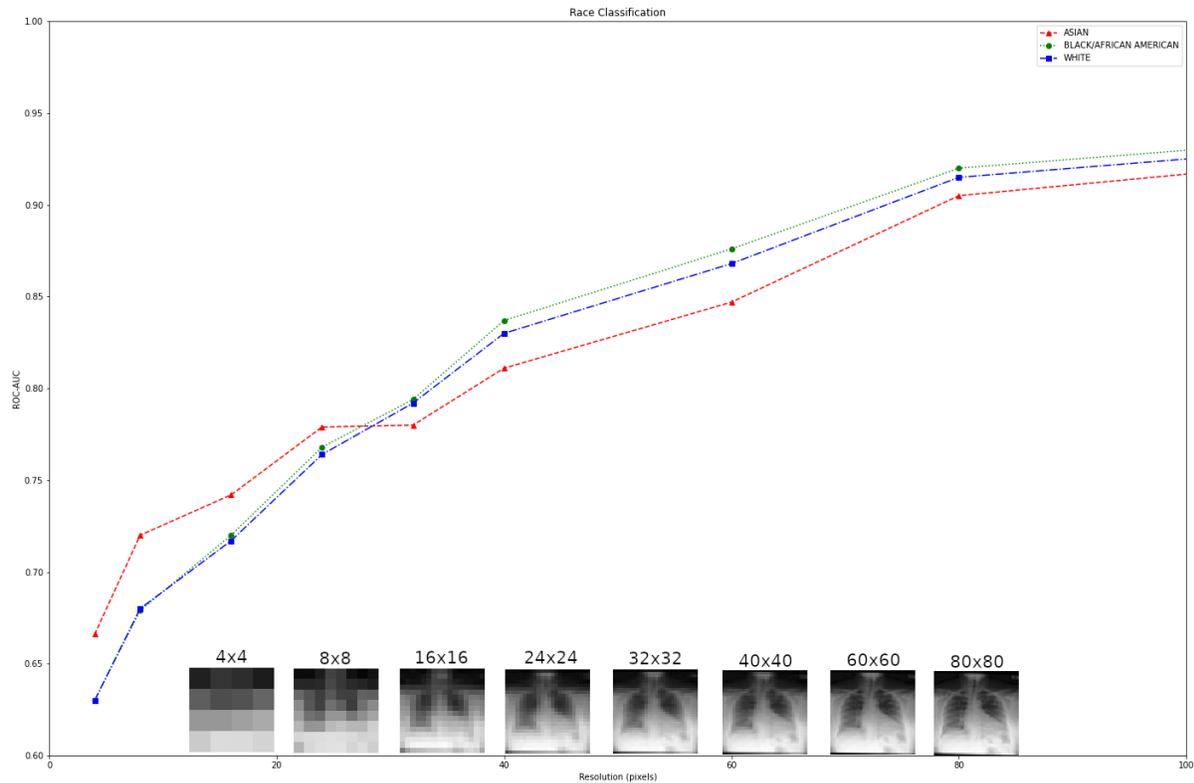

Supplemental Table 14 shows the AUC of race detection at various image resolutions from 4x4 resolution to 512x512 .

| Race | Resolution | | | | | | | | | | | |
|---|---|---|---|---|---|---|---|---|---|---|---|---|
| | 4 | 8 | 16 | 24 | 32 | 40 | 60 | 80 | 160 | 240 | 320 | 512 |
| Asian | 0.66 | 0.72 | 0.74 | 0.78 | 0.78 | 0.81 | 0.85 | 0.90 | 0.95 | 0.96 | 0.97 | 0.98 |
| Black | 0.63 | 0.68 | 0.72 | 0.77 | 0.79 | 0.84 | 0.88 | 0.92 | 0.96 | 0.97 | 0.97 | 0.98 |
| White | 0.63 | 0.68 | 0.72 | 0.76 | 0.79 | 0.83 | 0.87 | 0.92 | 0.96 | 0.96 | 0.97 | 0.97 |

Supplemental Table 15 shows AUCs for race detection using noisy and blurred CXR images on the MXR dataset. The AUCs of the noisy and blurred values show a drop in performance, although the AUCs are > 0.59 implying that some race information is still present in these images.

| | Asian | Black | White |
|---|---|---|---|
| MXR Densenet121-Original | 0.93 | 0.94 | 0.94 |
| MXR Densenet121-Noisy | 0.64 | 0.72 | 0.70 |
| MXR Densenet121-Blurred | 0.59 | 0.64 | 0.62 |

Supplemental Table 16 shows AUCs for race detection after masking the regions of interest indicated by the saliency maps. The AUCs decreased when the regions in the CXRs with the highest attention by the model were blocked out, but still maintained more than random chance of race detection.

| | Asian | Black | White |
|---|---|---|---|
| MXR Densenet121-Original | 0.93 | 0.94 | 0.94 |
| MXR Densenet121-Masked | 0.88 | 0.79 | 0.79 |

Supplemental Table 17 shows Comparative AUC values for the entire non segmented CXR, non lung and lung segmentations. Lung segmentations have the least AUC values while the original images have the highest AUCs. Race information is likely a combination of information from all portions of the image.

| | Asian | Black | White |
|---|---|---|---|
| MXR Densenet121-Original | 0.93 | 0.94 | 0.94 |
| MXR Densenet121-Non lung | 0.87 | 0.85 | 0.87 |
| MXR Densenet121-Lung | 0.68 | 0.74 | 0.73 |

Supplemental Table 18: AUROC performance of classifiers trained on 14 binary disease labels to predict race in MXR and CXP. Classifiers used are XGBoost (XGB), L1-regularized logistic regression (LR) and random forest (RF).

| | MXR | | | CXP | | |
|---|---|---|---|---|---|---|
| | XGB | LR | RF | XGB | LR | RF |
| **White** | 57.1% | 56.9% | 56.9% | 52.1% | 51.9% | 51.9% |
| **Black** | 60.8% | 60.6% | 60.5% | 56.9% | 56.6% | 56.8% |
| **Asian** | 56.1% | 54.8% | 56.8% | 54.3% | 54.2% | 54.2% |

| | | Asian | Black | White |
|---|---|---|---|---|
| | Supplemental Figure 5 : Saliency maps for primary race detection for CXR images. The saliency maps were assessed qualitatively by all group members, across all tasks, including by members with radiology expertise. No consistent anatomical localisation was appreciated, and no anatomic structures appeared to be particularly salient to the decision making process. | | | |
| A1 | Accurate primary race prediction | 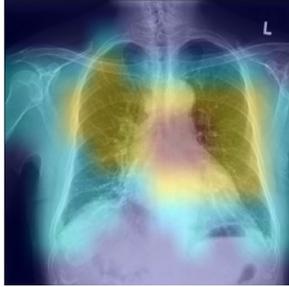 | 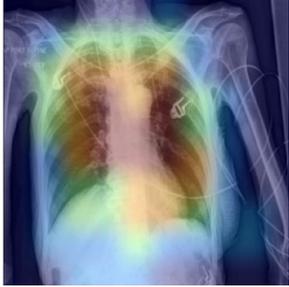 | 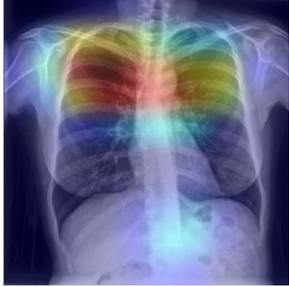 |
| B3 | Accurate primary race prediction from the "no finding" class label | 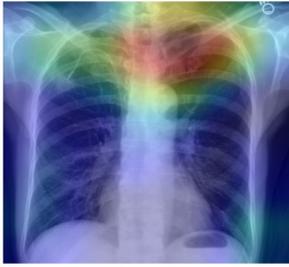 | 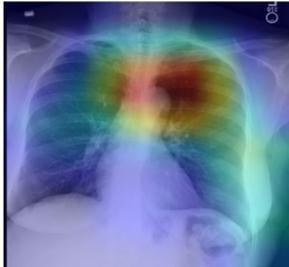 | 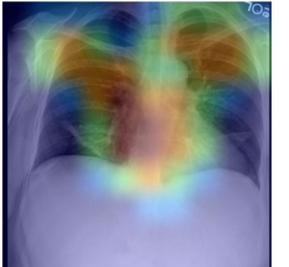 |
| | | 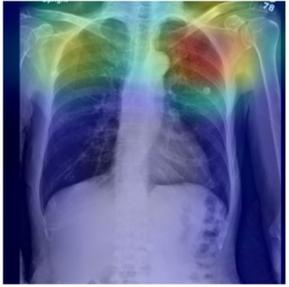 | 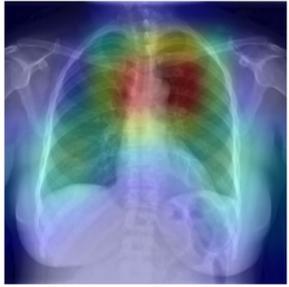 | 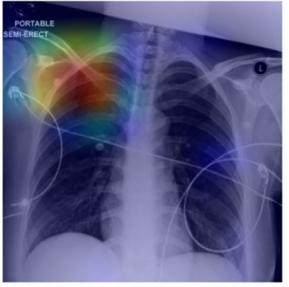 |
| A1 | Incorrectly classified race prediction | **Incorrectly predicted Asian** 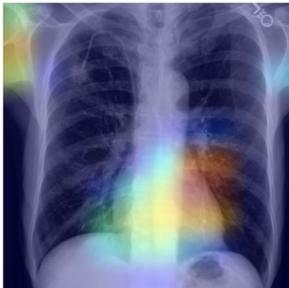 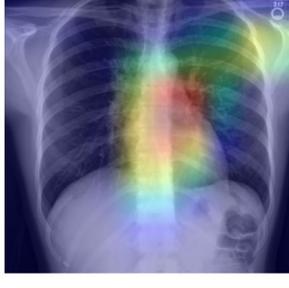 | **Incorrectly predicted Black** 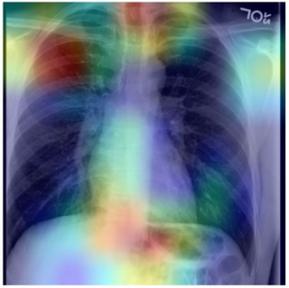 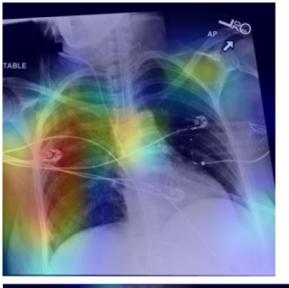 | **Incorrectly predicted White** 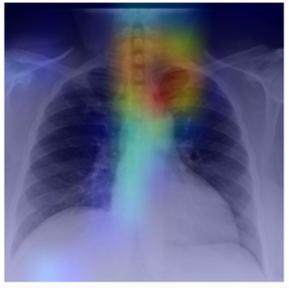 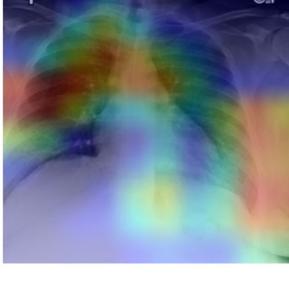 |

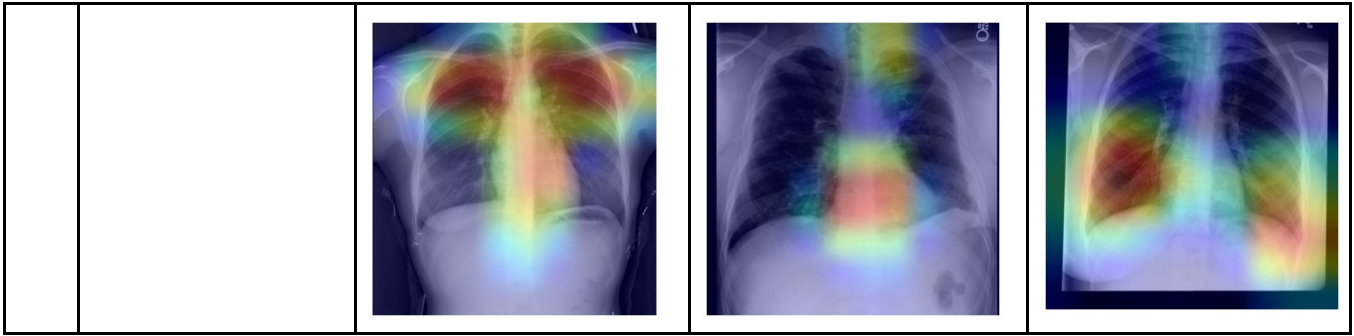

Supplemental Figure 6 shows generated saliency maps for Black and White patients across various breast density classes. Visual assessment by the team including four radiologists of a random sample of saliency maps did not produce any identifiable pattern that could explain race prediction.

| Race/Breast Density | 1 (Fatty) | 2 (Scattered) | 3 (Heterogeneous) | 4 (Dense) |
|---|---|---|---|---|
| Original Image | 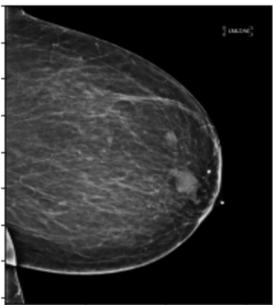 | 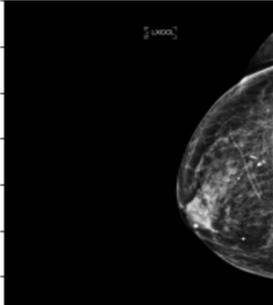 | 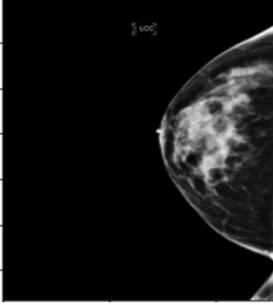 | 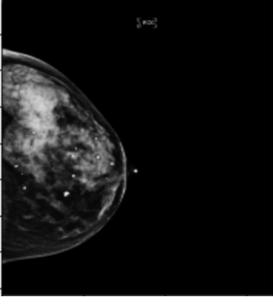 |
| Black | 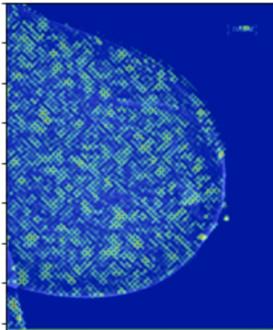 | 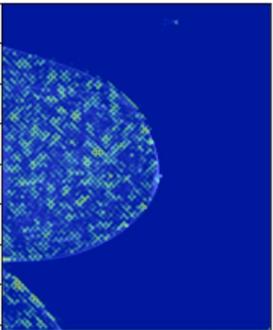 | 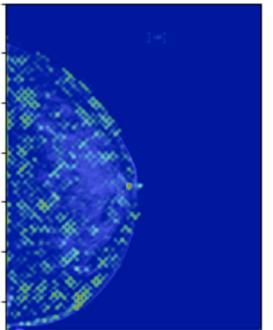 | 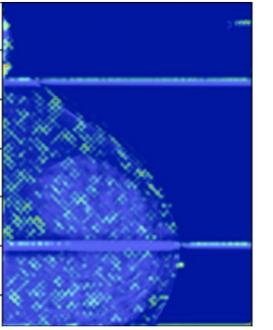 |
| White | 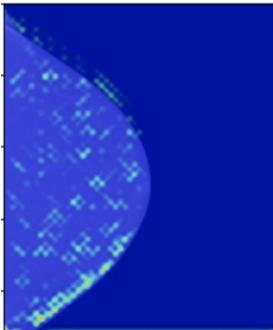 | 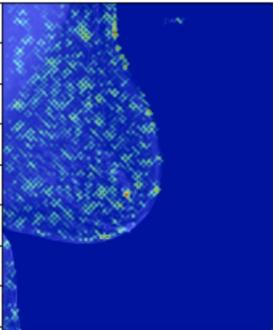 | 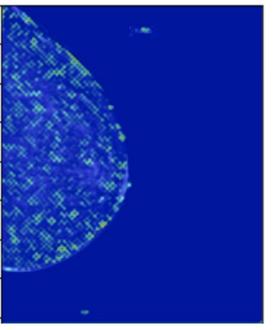 | 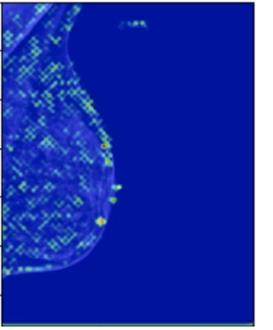 |

Supplemental Figure 7 : Saliency maps for primary race detection for non CXR images. The saliency maps were assessed qualitatively by all group members, across all tasks, including by members with radiology expertise. No consistent anatomical localisation was appreciated, and no anatomic structures appeared to be particularly salient to the decision making process.

| A2 | Accurate primary race prediction - Digital Hand Atlas | Black | White |
|---|---|---|---|
| | | 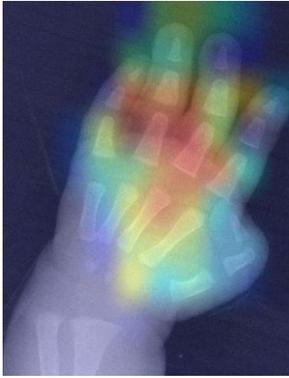 | 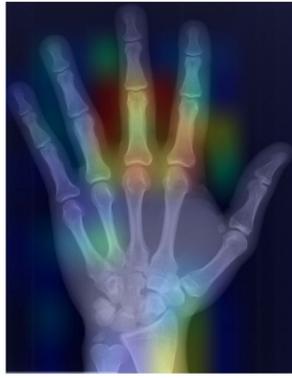 |
| | Incorrectly classified race prediction - DHA | Black incorrectly classified as White | White incorrectly classified as Black |
| | | 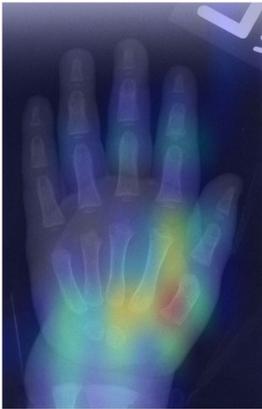 | 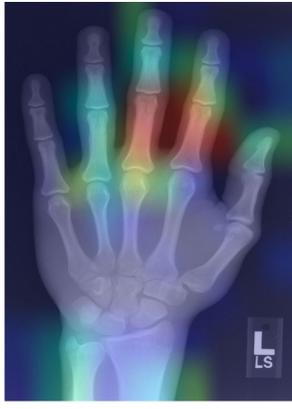 |
| A2 | Accurate primary race prediction - CT Chest | Black | White |
| | | 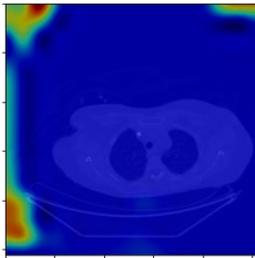 | 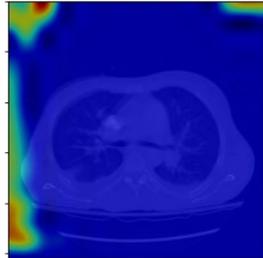 |
| | Incorrectly classified race | Black incorrectly classified as White | White incorrectly classified as Black |
| | prediction - CT Chest | 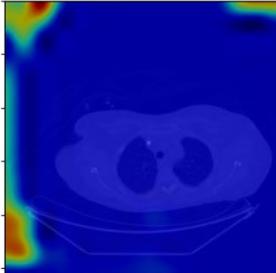 | 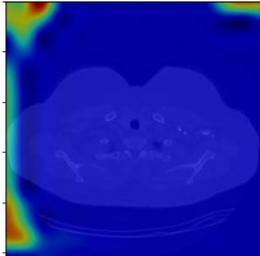 |
| A2 | Accurate race prediction - C Spine radiographs | Black<br>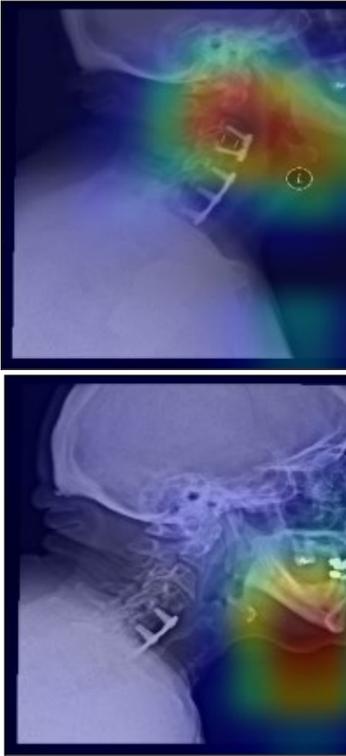 | White<br>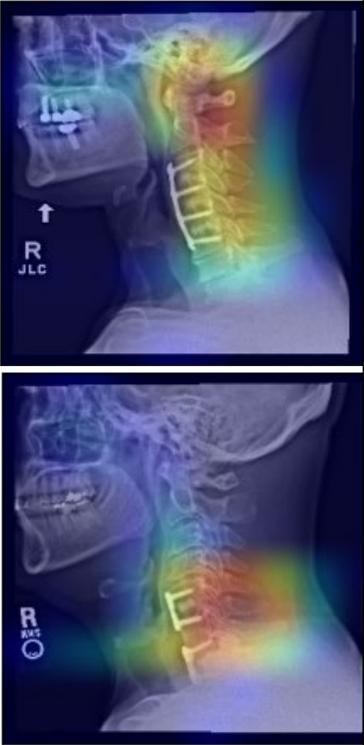 |
| | Incorrectly classified race prediction - C Spine radiographs | Black incorrectly classified as White<br>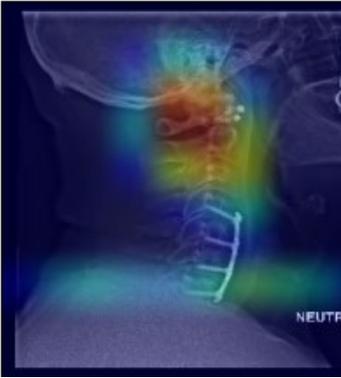 | White incorrectly classified as black<br>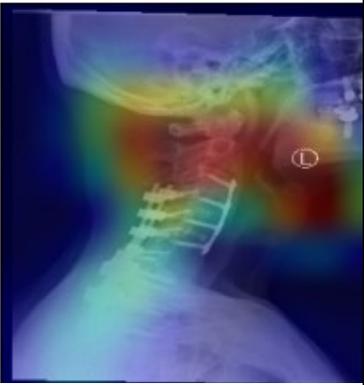 |

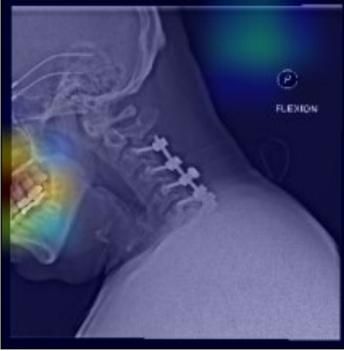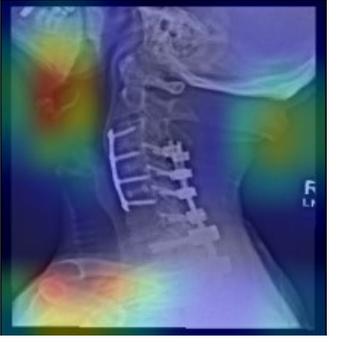

**Supplementary References**